%% file: main_paper.tex
\documentclass{llncs}
\usepackage{makeidx}  
\usepackage{paralist} 
\usepackage{amsmath}
\usepackage{amssymb}
\usepackage{graphicx}
\usepackage{multirow}
\usepackage{color}
\usepackage{wrapfig}
\usepackage[hidelinks]{hyperref}
\usepackage{caption,subcaption}
\usepackage{booktabs}
\usepackage{paralist}
\usepackage[]{changes}

\usepackage{textpos}

\DeclareMathOperator*{\argmax}{arg\,max}
\definechangesauthor[name={Satya Almasian},color=red]{sa}

\begin{document}

\title{Word Embeddings for Entity-annotated Texts}
\titlerunning{Word Embeddings for Entity-annotated Texts}  
%
\author{Satya Almasian \and Andreas Spitz \and Michael Gertz}
\institute{Heidelberg University, Heidelberg, Germany\\
\email{\{almasian, spitz, gertz\}@informatik.uni-heidelberg.de}}


\maketitle              


\input{section_0_abstract.tex}

\section{Introduction}
\label{sec:intro}
\input{section_1_introduction.tex}

\section{Related Work}
\label{sec:related}
\input{section_2_relatedwork.tex}

\section{Embedding Models}
\label{sec:model}
\input{section_3_model.tex}

\section{Evaluation of Embeddings}
\label{sec:topic}
\input{section_4_eval_basic.tex}

\section{Experimental Exploration of Entity Embeddings}
\label{sec:exp}
\input{section_5_experimental.tex}

\section{Conclusion and Ongoing Work}
\label{sec:conclusion}
\input{section_6_conclusion.tex}


\bibliographystyle{splncs03}
\bibliography{bibliography} 

\end{document}

%% file: section_0_abstract.tex
\begin{abstract}
Learned vector representations of words are useful tools for many information retrieval and natural language processing tasks due to their ability to capture lexical semantics. However, while many such tasks involve or even rely on named entities as central components, popular word embedding models have so far failed to include entities as first-class citizens. While it seems intuitive that annotating named entities in the training corpus should result in more intelligent word features for downstream tasks, performance issues arise when popular embedding approaches are na\"{i}vely applied to entity annotated corpora. Not only are the resulting entity embeddings less useful than expected, but one also finds that the performance of the non-entity word embeddings degrades in comparison to those trained on the raw, unannotated corpus.
In this paper, we investigate approaches to jointly train word and entity embeddings on a large corpus with automatically annotated and linked entities. We discuss two distinct approaches to the generation of such embeddings, namely the training of state-of-the-art embeddings on raw-text and annotated versions of the corpus, as well as node embeddings of a co-occurrence graph representation of the annotated corpus.
We compare the performance of annotated embeddings and classical word embeddings on a variety of word similarity, analogy, and clustering evaluation tasks, and investigate their performance in entity-specific tasks. Our findings show that it takes more than training popular word embedding models on an annotated corpus to create entity embeddings with acceptable performance on common test cases. Based on these results, we discuss how and when node embeddings of the co-occurrence graph representation of the text can restore the performance.
\keywords{word embeddings, entity embeddings, entity graph}
\end{abstract}

%% file: section_1_introduction.tex

Word embeddings are methods that represent words in a continuous vector space by mapping semantically similar or related words to nearby points. These vectors can be used as features in NLP or information retrieval tasks, such as query expansion~\cite{DBLP:journals/corr/DiazMC16,Kuzi:2016:QEU:2983323.2983876}, named entity recognition~\cite{Das:2017:NER:3041821.3015467}, or document classification~\cite{R17-1057}. The current style of word embeddings dates back to the neural probabilistic model published by Bengio et al.~\cite{DBLP:conf/nips/BengioDV00}, prior to which embeddings were predominantly generated by latent semantic analysis methods~\cite{Deerwester90indexingby}. However, most developments are more recent. The two most popular methods are word2vec proposed by Mikolov et al.~\cite{Mikolov}, and GloVe by Pennington et al.~\cite{DBLP:conf/emnlp/PenningtonSM14}. Since then, numerous alternatives to these models have been proposed, often for specific tasks.

Common to all the above approaches is an equal treatment of words, without word type discrimination. Some effort has been directed towards embedding entire phrases~\cite{DBLP:journals/tacl/HillCKB16,DBLP:conf/acl/YinS14} or combining compound words after training~\cite{DBLP:conf/eacl/DurmeRPM17,DBLP:journals/cogsci/MitchellL10}, but entities are typically disregarded, which entails muddied embeddings with ambiguous entity semantics as output. 
For example, an embedding that is trained on ambiguous input is unable to distinguish between instances of \emph{Paris}, which might refer to the French capital, the American heiress, or even the Trojan prince. Even worse, entities can be conflated with homographic words, e.g., the former U.S. president \emph{Bush}, who not only shares ambiguity with his family members, but also with shrubbery. Moreover, word embeddings are ill-equipped to handle synonymous mentions of distinct entity labels without an extensive local clustering of the neighbors around known entity labels in the embedding space.

Joint word and entity embeddings have been studied only for task-specific applications, such as entity linkage or knowledge graph completion. Yamada et al.~\cite{DBLP:conf/conll/YamadaS0T16} and Moreno et al.~\cite{DBLP:conf/esws/MorenoBBDLRTG17} propose entity and word embedding models specific to named entity linkage by using knowledge bases. Embedding entities in a knowledge graph has also been studied for relational fact extraction and knowledge base completion~\cite{DBLP:conf/emnlp/ToutanovaCPPCG15,DBLP:conf/emnlp/WangZFC14}. All of these methods depend on knowledge bases as an external
source of information, and often train entity and
word embeddings separately and combine them afterwards. However, it seems reasonable to avoid
this separation and learn embeddings
directly from annotated text to create general-purpose entity embeddings, jointly with word embeddings.

Typically, state-of-the-art entity recognition and linking is dependent on an extensive NLP-stack that includes sentence splitting, tokenization, part-of-speech tagging, and entity recognition, with all of their accrued cumulative errors. Thus, while embeddings stand to benefit from the annotation and resolution of entity mentions, an analysis of the drawbacks and potential applications is required.
In this paper, we address this question by using popular word embedding methods to jointly learn word and entity embeddings from an automatically annotated corpus of news articles. We also use cooccurrence graph embeddings as an alternative, and rigorously evaluate these for a comprehensive set of evaluation tasks. Furthermore, we explore the properties of  
our models in comparison to plain word embeddings to estimate their usefulness for entity-centric tasks.

\noindent
\textbf{Contributions.}
In the following, we make five contributions.
\begin{inparaenum}[$(i)$]
\item We investigate the performance of popular word embedding methods when trained on an entity-annotated corpus, and 
\item introduce graph-based node embeddings as an alternative that is trained on a cooccurrence graph representations of the annotated text\footnote{Source code available at: \url{https://github.com/satya77/Entity_Embedding}}. 
\item We compare all entity-based models to traditional word embeddings on a comprehensive set of word-centric intrinsic evaluation tasks, and introduce entity-centric intrinsic tasks.
\item We explore the underlying semantics of the embeddings and implications for entity-centric downstream applications, and 
\item discuss the advantages and drawbacks of the different models.
\end{inparaenum}

%% file: section_2_relatedwork.tex

Related work covers word and graph embeddings, as well as cooccurrence graphs.

\noindent
\textbf{Word embeddings.} 
A word embedding, defined as a mapping $V\rightarrow \mathbb{R}^{d}$, maps a word $w$ from a vocabulary $V$ to a vector $\theta$ in a $d$-dimensional embedding space~\cite{schnabel2015evaluation}. To learn such embeddings, window-based models employ supervised learning, where the objective is to predict a word's context when given a center word in a fixed window. Mikolov et al.\ introduced the continuous bag-of-words (CBOW) and the skip-gram architecture as  window-based models that are often referred to as word2vec~\cite{Mikolov}. The CBOW architecture predicts the current word based on the context, while skip-gram predicts surrounding words given the current word. This model was later improved by Bojanowski et al.\ to take character level information into account~\cite{DBLP:journals/corr/BojanowskiGJM16}. In contrast to window-based models, matrix factorization methods operate directly on the word cooccurrence matrix. Levy and Goldberg showed that implicitly factorizing a word-context matrix, whose cells contain the point-wise mutual information of the respective word and context pairs, can generate embeddings close to word2vec~\cite{NIPS2014_5477}. Finally, the global vector model (GloVe) combines the two approaches and learns word embeddings by minimizing the distance between the number of cooccurrences of words and the dot product of their vector representations~\cite{DBLP:conf/emnlp/PenningtonSM14}.

\noindent
\textbf{Graph node embeddings.}
A square word cooccurrence matrix can be interpreted as a graph whose nodes correspond the rows and columns, while the matrix entries indicate edges between pairs of nodes. The edges can have weights, which usually reflect some distance measure between the words, such as the number of tokens between them. These networks are widely used in natural language processing, for example in summarization~\cite{W04-3252} or word sense discrimination~\cite{Ferret:2004:DWS:1220355.1220549}. More recent approaches have included entities in graphs to support information retrieval tasks, such as topic modeling~\cite{DBLP:conf/ecir/SpitzG18}. In a graph representation of the text, the neighbors of a node can be treated as the node's context. Thus, embedding the nodes of a graph also results in embeddings of words.

Numerous node embedding techniques for graph nodes exist, which differ primarily in the similarity measure that is used to define node similarity. DeepWalk was the first model to learn latent representations of graph nodes by using sequences of fixed-length random walks around each node~\cite{Perozzi:2014:DOL:2623330.2623732}. Node2vec improved the DeepWalk model by proposing a flexible neighborhood sampling strategy that interpolates between depth-first and breadth-first search~\cite{Node2vec}. The LINE model learns a two-part embedding, where the first part corresponds to the first-order proximity (i.e., the local pairwise proximity between two vertices) and the second part represents the second-order proximity (i.e., the similarity between their neighborhood structures)~\cite{tang2015line}. More recently, Tsitsulin et al.\ proposed VERSE, which supports multiple similarity functions that can be tailored to individual graph structures~\cite{VERSE}. With VERSE, the user can choose to emphasize the structural similarity or focus on an adjacency matrix, thus emulating the first-order proximity of LINE. 
Due to this versatility, we thus focus on DeepWalk and VERSE as representative node embedding methods to generate joint entity and word embeddings from cooccurrence graphs.

%% file: section_3_model.tex

To jointly embed words and named entities, we tweak existing word and graph node embedding techniques. To na\"{i}vely include entities in the embeddings, we train the state-of-the-art word embedding methods on an entity annotated corpus. As an alternative, we transform the text into a cooccurrence graph and use graph-based models to train node embeddings. We compare both models against models trained on the raw (unannotated) corpus. In this section, we first give an overview of GloVe and word2vec for raw text input. Second, we describe how to include entity annotations for these models. Finally, we show how DeepWalk and VERSE can be used to obtain entity embeddings from a cooccurrence graph. 

\subsection{Word Embeddings on Raw Text }
State-of-the-art word embedding models are typically trained on the raw text that is cleaned by removing punctuation and stop words. Since entities are not annotated, all words are considered as terms. We use skip-gram from word2vec ($r$W2V), and the GloVe model ($r$GLV), where $r$ denotes raw text input.

\noindent
 \textbf{Skip-gram} aims to optimize the embeddings $\theta$, which maximize the corpus probability over all words $w$ and their contexts $c$ in documents $D$~\cite{DBLP:journals/corr/GoldbergL14} as 
\begin{equation}
\argmax_{\theta} \prod_{(w,c)\in D} p(c \mid w, \theta )
\label{eq:skip-gram}
\end{equation}
To find the optimal value of $\theta$, the conditional probability is modelled using softmax and solved with negative sampling or hierarchical softmax.

\noindent
\textbf{GloVe} learns word vectors such that their dot product equals the logarithm of the words' cooccurrence probability~\cite{DBLP:conf/emnlp/PenningtonSM14}. If $X\in \mathbb{R}^{ W\times W }$ is a matrix of word cooccurrence counts $X_{ij}$, then GloVe optimizes embeddings $\theta_i$ and $\tilde{ \theta_j }$ for center words $i$ and context words $j$, and biases $b$ and $\tilde{b}$ to minimize the cost function 
\begin{equation}
J=\sum _{ i,j=1 }^{ W }{ f({ X }_{ ij } } )(\theta_{ i }^{ T }\tilde{  \theta_{ j } } +b_{ i }+\tilde{  b_{ j } } -log{ X }_{ ij })^2, \quad
f=\left\{
  \begin{array}{@{}ll@{}}
    (\frac { x }{ { x }_{ max } } )^{ \alpha  }& \text{if}\ x<x_{max} \\
    1 & \text{otherwise}
  \end{array}\right.
\label{eq:glove_cost}
\end{equation}
The function $f$ serves as an upper bound on the maximum number of allowed word cooccurrences $x_{max}$, with $\alpha \in [0,1]$ as an exponential dampening factor.

\subsection{Word Embeddings on Annotated Text}
Named entities are typically mentions of person, organization, or location names, and numeric expressions, such as dates or monetary values in a text~\cite{Nadeau}. Formally, if $T$ denotes the set of terms in the vocabulary (i.e, words and multi-word expressions), let $N\subseteq T$ be the subset of named entities. Identifying these mentions is a central problem in natural language processing that involves part-of-speech tagging, named entity recognition, and disambiguation. Note that $T$ is technically a multi-set since multiple entities may share ambiguous labels, but entities can be represented by unique identifiers in practice. Since annotated texts contain more information and are less ambiguous, embeddings trained on such texts thus stand to perform better in downstream applications.
To generate these embeddings directly, we use word2vec and GloVe on a corpus with named entity annotations and refer to them as $a$W2V and $a$GLV, where $a$ denotes the use of annotated text. Since entity annotation requires part-of-speech tagging, we use POS tags to remove punctuation and stop word classes. Named entity mentions are identified and replaced with unique entity identifiers. The remaining words constitute the set of terms $T\setminus N$ and are used to generate term cooccurrence counts for the word embedding methods described above. 

\subsection{Node Embeddings of Cooccurrence Graphs}
A cooccurrence graph $G=(T,E)$ consists of a set of terms $T$ as nodes and a set of edges $E$ that connect cooccurring terms. Edges can be weighted, where the weights typically encode some form of textual distance or similarity between the terms. If the graph is extracted from an annotated corpus, some nodes represent named entities. For entity annotations in particular, implicit networks can serve as graph representations that use similarity-based weights derived from larger cross-sentence cooccurrences of entity mentions~\cite{Spitz2016a}. By embedding nodes in these networks, we also obtain embeddings of both entities and terms.

From the available node embedding methods, we select a representative subset. While it is popular, we omit node2vec since cooccurrence graphs are both large and dense, and node2vec tends to be quite inefficient for such graphs~\cite{DBLP:journals/corr/abs-1805-00280}. Similarly, we do not use LINE since the weighted cooccurrence graphs tend to have an unbalanced distribution of frequent and rare words, meaning that the second-order proximity of LINE becomes ill-defined. Since the adjacency similarity of VERSE correlates with the first-order proximity in LINE, we use VERSE (VRS) as a representative of the first-order proximity and DeepWalk (DW) as a representative of random walk-based models. Conceptually, graph node embeddings primarily differ from word embeddings in the sampling of the context.

\noindent
\textbf{DeepWalk} performs a series of fixed-length random walks on the graph to learn a set of parameters $\Theta_E \in R^{T\times d} $, where $d$ is a small number of latent dimensions. The nodes visited in a random walk are considered as context and are used to train a skip-gram model. DeepWalk thus maximizes the probability of observing the $k$ previous and next nodes in a random walk starting at node $t_{i}$ by minimizing the negative logarithmic probability to learn the node embedding $\theta$~\cite{DBLP:journals/corr/GoyalF17}: 
\begin{equation}
J=-\log P( t_{i-k},...,t_{i-1},t_{i+1},...,t_{i+k} \mid \theta )
\label{eq:dw}
\end{equation}
Since cooccurrence graphs are weighted, we introduce weighted random walks that employ a transition probability to replace the uniform random walks. The probability of visiting node $j$ from node $i$ is then proportional to the edge weight $e_{i,j}$, where $ E_{ i }$ denotes the set of all edges starting at node $t_{i}$
\begin{equation}
P_{i,j}= \frac{f(e_{i,j})}{\sum_{ e_{ik} \in E_i } f(e_{ik}) }
\label{eq:edge_weight}
\end{equation}
and $f$ is a normalization function. To create a more balanced weight distribution, we consider no normalization, i.e., $f=\mathrm{id}$, and a logarithmic normalization, i.e., $f=\log$. We refer to these as (DW$_{id}$) and (DW$_{log}$), respectively. The performance of $f=\mathrm{sqr}$ is similar to a logarithmic normalization and is omitted.

\noindent
\textbf{VERSE} is designed to accept any node similarity measure for context selection~\cite{VERSE}. Three measures are part of the original implementation, namely Personalized PageRank, adjacency similarity, and SimRank. 
SimRank is a measure of structural relatedness and thus ill-suited for word relations. 
Personalized Page\-Rank is based on the stationary distribution of a random walk with restart, and essentially replicates DeepWalk. Thus, we focus on adjacency similarity, which derives node similarities from the outgoing degree $out(t_{i})$ of node $t_{i}$: 

\begin{equation}
sim^{ ADJ }_{ G }(t_{ i },t_{ j })=\left\{ \begin{matrix} \frac { 1 }{ out(t_{ i }) } \quad if\quad (t_{ i },t_{ j })\in E\quad  \\ 0\quad \qquad \qquad otherwise \end{matrix} \right.
\label{eq:adj}
\end{equation}
The model then minimizes the Kullback-Leibler divergence between the similarity measure of two nodes and the dot product of their embeddings $\theta_{i}$ and $\theta_{j}$, and thus works conceptually similar to GloVe. In the following, we use this model as our second node embedding approach and refer to it as VRS.

%% file: section_4_eval_basic.tex
In the following, we look at the datasets used for training and evaluation, before comparing the learned models on typical tasks and discussing the results.

\subsection{Evaluation Tasks}
The main benefit of word embeddings is found in downstream applications (extrinsic evaluation). 
However, since these evaluations are task-specific, an embedding that works well for one task may fail for another. The more common test scenario is thus intrinsic and analyzes how well the embeddings capture syntactic or semantic relations~\cite{schnabel2015evaluation}. The problem with such tests is that the notion of semantics is not universal~\cite{DBLP:journals/corr/abs-1801-09536}. Some datasets reflect semantic relatedness and some semantic similarity~\cite{Klob}. 
Since few intrinsic datasets include entities, we focus on the performance of term-based intrinsic tasks. Following the approach by Schnabel et al.~\cite{schnabel2015evaluation}, we use three kinds of intrinsic evaluations.

\noindent
\textbf{Relatedness} uses datasets with relatedness scores for pairs of words annotated by humans. The cosine similarity or Euclidean distance between the embeddings of two words should have a high correlation with scores assigned by humans. 
\begin{enumerate}[$i)$]
 \item \emph{Similarity353:} $203$ instances of similar word pairs from WordSim353~\cite{DBLP:conf/naacl/AgirreAHKPS09} classified as synonyms, antonyms, identical, and unrelated pairs~\cite{Agirre:2009:SSR:1620754.1620758}.
 \item \emph{Relatedness353:} $252$ instances of word pairs from WordSim353~\cite{DBLP:conf/naacl/AgirreAHKPS09} that are not similar but still considered related by humans, and unrelated pairs~\cite{Agirre:2009:SSR:1620754.1620758}.
 \item \emph{MEN:} $3,000$ word pairs with human-assigned similarity judgements~\cite{Bruni:2014:MDS:2655713.2655714}.
 \item \emph{RG65:} $65$ pairs with annotated similarity, scaling from $0$ to $4$~\cite{Rubenstein:1965:CCS:365628.365657}.
 \item \emph{RareWord:} $2,034$ rare word pairs with human-assigned similarity scores~\cite{W13-3512}.
 \item \emph{SimLex-999:} $999$ pairs of human-labeled examples of semantic relatedness~\cite{DBLP:journals/corr/HillRK14}.
 \item \emph{MTurk:} $771$ words pairs with semantic relatedness scores from
$0$ to $5$~\cite{Radinsky:2011:WTC:1963405.1963455}.
\end{enumerate}

\noindent
\textbf{Analogy.} In the analogy task, the objective is to find a word $y$ for a given word $x$, such that  $x : y$ best resembles a sample relationship $a : b$. Given the triple $(a,b,x)$ and a target word $y$, the nearest neighbour of $\hat{\theta}:=\theta_a -\theta_b + \theta_x$ is computed and compared to $y$. If $y$ is the word with the highest cosine similarity to $\hat{\theta}$, the task is solved correctly. 
For entity embeddings, we can also consider an easier, type-specific variation of this task, which only considers neighbors that match a given entity class, such as locations or persons. 
\begin{enumerate}[$i)$]
  \item \emph{GA}: The \emph{Google Analogy} data consists of $19,544$ morphological and semantic questions used in the original word2vec publication~\cite{Mikolov}. Beyond terms, it contains some location entities that support term to city relations. 
  \item \emph{MSR:} The Microsoft Research Syntactic analogies dataset contains $8,000$ morphological questions~\cite{mikolov2013linguistic}. All word pairs are terms.
\end{enumerate}

\noindent
\textbf{Categorization.} When projecting the embeddings to a $2$- or $3$-dimensional space with t-SNE~\cite{vanDerMaaten2008} or principle component analysis~\cite{Abdi:2010:PCA:3160436.3160440}, we expect similar words to form meaningful clusters, which we can evaluate by computing the purity of clusters~\cite{schutze2008introduction}. We use two datasets from the Lexical
Semantics Workshop, which do not contain entities. Additionally, we create three datasets by using Wikidata to find entities of type person, location, and organization. 
\begin{enumerate}[$i)$]
  \item \emph{ESSLLI\_1a:} $44$ concrete nouns that belong to six semantic categories~\cite{Baroni}. 
  \item \emph{ESSLLI\_2c:} $45$ verbs that belong to nine semantic classes~\cite{Baroni}.
  \item \emph{Cities:} $150$ major cities in the U.S., the U.K., and Germany. 
  \item \emph{Politicians:} $150$ politicians from the U.S., the U.K., and Germany.
  \item \emph{Companies:} $110$ software companies, Web services, and car manufacturers. 
\end{enumerate}

\subsection{Training Data}\label{training_data}
For training, we use $209,023$ news articles from English-speaking news outlets, collected from June to November $2016$ by Spitz and Gertz~\cite{DBLP:conf/ecir/SpitzG18}. The data contains a total of $5,427,383$ sentences. To train the regular word embeddings, we use the raw article texts, from which we remove stop words and punctuation. For the annotated embeddings, we extract named entities with Ambiverse\footnote{\url{https://github.com/ambiverse-nlu}}, a state-of-the-art annotator that links entity mentions of persons, locations, and organizations to Wikidata identifiers. Temporal expressions of type date are annotated and normalized with HeidelTime~\cite{DBLP:journals/lre/StrotgenG13}, and part-of-speech annotations are obtained from the Stanford POS tagger~\cite{DBLP:conf/naacl/ToutanovaKMS03}. We use POS tags to remove punctuation and stop words (wh-determiner, pronouns, auxiliary verbs, predeterminers, possessive endings, and prepositions).
To generate input for the graph-based embeddings, we use the extraction code of the LOAD model~\cite{Spitz2016a} that generates implicit weighted graphs of locations, organizations, persons, dates, and terms, where the weights encode the textual distance between terms and entities that cooccur in the text. We include term cooccurrences only inside sentences and entity-entity cooccurrences up to a default window size of five sentences. The final graph has $T=93,390$ nodes (terms and entities) and $E=9,584,191$ edges. Since the evaluation datasets contain words that are not present in the training vocabulary, each data set is filtered accordingly.

\subsection{Parameter Tuning}\label{param}
We perform extensive parameter tuning for each model and only report the settings that result in the best performance. Since the embedding dimensions have no effect on the relative difference in performance between models, all embeddings have $100$ dimensions. Due to the random initialization at the beginning of the training, all models are trained $10$ times and the performance is averaged.

\noindent
\textbf{Word2vec-based models} 
are trained with a learning rate of $0.015$ and a window size of $10$. We use $8$ negative samples on the raw data, and $16$ on the annotated data. Words with a frequency of less than $3$ are removed from the vocabulary as there is not enough data to learn a meaningful representation.

\noindent
\textbf{GloVe-based models} are trained with a learning rate of $0.06$. For the weighting function, a scaling factor of $0.5$ is used with a maximum cut-off of $1000$. Words that occur less than $5$ times are removed from the input.

\noindent
\textbf{DeepWalk models} use $100$ random walks of length $4$ from each node. Since the cooccurrence graph has a relatively small diameter, longer walks would introduce unrelated words into contexts. We use a learning rate of $0.015$ and $64$ negative samples for the skip-gram model that is trained on the random walk results.

\noindent
\textbf{VERSE models} use a learning rate of $0.025$ and $16$ negative samples.

A central challenge in the comparison of the models is the fact that the training process of graph-based and textual methods is incomparable. On the one hand, the textual models consider one pass through the corpus as one iteration. On the other hand, an increase in the number of random walks in DeepWalk increases both the performance and the runtime of the model, as it provides more data for the skip-gram model. In contrast, the VERSE model has no notion of iteration and samples nodes for positive and negative observations.
To approach a fair evaluation, we thus use similar training times for all models (roughly $10$ hours per model on a $100$ core machine). We fix the number of iterations of the textual models and DeepWalk's skip-gram at $100$. For VERSE, we use $50,000$ sampling steps to obtain a comparable runtime.

\subsection{Evaluation Results}
Unsurprisingly, we find that no single model performs best for all tasks. The results of the relatedness task are shown in Table~\ref{wordsim_normal}, which shows that word2vec performs better than GloVe with this training data. The performance of both methods degrades slightly but consistently when they are trained on the annotated data in comparison to the raw data. The DeepWalk-based models perform better than GloVe but do poorly overall. VERSE performs very well for some of the tasks, but is worse than word2vec trained on the raw data for rare words and the SimLex data. This is likely caused by the conceptual structure of the cooccurrence graph on which VERSE is trained, which captures relatedness and not similarity as tested by SimLex. For the purely term-based tasks in this evaluation that do not contain entity relations, word2vec is thus clearly the best choice for similarity tasks, while VERSE does well on relatedness tasks.

\begin{table}[t]
\caption{Word similarity results. Shown are the Pearson correlations between the cosine similarity of the embeddings and the human score on the word similarity datasets. The two best values per task are highlighted.}
\label{wordsim_normal}
\setlength{\tabcolsep}{6pt} 
\renewcommand{\arraystretch}{1.0} 
\resizebox{\textwidth}{!}{%
\begin{tabular}{lcccccccc}
\toprule
 & $r$W2V & $r$GLV & $a$W2V & $a$GLV & DW$_{id}$ & DW$_{log}$ & VRS \\
\midrule
Similarity353      & \textbf{0.700}   & 0.497   & \textbf{0.697} & 0.450 & 0.571  & 0.572     & 0.641\\
Relatedness353     & \textbf{0.509}   & 0.430   & 0.507& 0.428& 0.502  & 0.506   & \textbf{0.608}\\
MEN                & \textbf{0.619}   & 0.471   & \textbf{0.619}& 0.469& 0.539  & 0.546     & \textbf{0.640}\\ 
RG65               & \textbf{0.477}   & 0.399   & 0.476& 0.386& 0.312  & 0.344     & \textbf{0.484}\\ 
RareWord           & \textbf{0.409}   & 0.276   & \textbf{0.409}& 0.274& \textbf{0.279}  & 0.276    & 0.205\\  
SimLex-999         & \textbf{0.319}   & 0.211   & \textbf{0.319}& 0.211& \textbf{0.279}  & 0.201      & 0.236\\ 
MTurk              & \textbf{0.647}   & 0.493   & 0.644& 0.502& 0.592  & 0.591   & \textbf{0.687}\\
\midrule
average            & \textbf{0.526}   & 0.400   & \textbf{0.524} & 0.389 & 0.439  & 0.433   & 0.500 \\
\bottomrule
\end{tabular}%
}
\vspace*{-10pt}
\end{table}

\begin{table}[t]
\caption{Word analogy results. Shown are the prediction accuracy for the normal analogy tasks and the variation in which predictions are limited to the correct entity type. The best two values per task and variation are highlighted.}
\label{analogy_normal}
\setlength{\tabcolsep}{2pt} 
\renewcommand{\arraystretch}{1.0} 
\resizebox{\textwidth}{!}{%
\begin{tabular}{lccccccccccccc}
\toprule
 & \multicolumn{7}{c}{normal analogy} && \multicolumn{5}{c}{typed analogy} \\
 \cmidrule(lr){2-8}
 \cmidrule(lr){10-14}
 & $r$W2V & $r$GLV & $a$W2V & $a$GLV & DW$_{id}$ & DW$_{log}$ & VRS 
   && $a$W2V & $a$GLV & DW$_{id}$ & DW$_{log}$ & VRS \\
\cmidrule(lr){1-1}
\cmidrule(lr){2-8}
\cmidrule(lr){10-14}
GA        & 0.013 & \textbf{0.019} & 0.003 & 0.015 & 0.009 & 0.009 & \textbf{0.035} 
   && 0.003 & \textbf{0.016} & 0.011 & 0.011 & \textbf{0.047} \\
MSR       & \textbf{0.014} & \textbf{0.019} & 0.001 & \textbf{0.014} & 0.002 & 0.002 & 0.012 
   && 0.001 & \textbf{0.014} & 0.002 & 0.002 & \textbf{0.012} \\
\cmidrule(lr){1-1}
\cmidrule(lr){2-8}
\cmidrule(lr){10-14}
avg       & 0.013 & \textbf{0.019} & 0.002 & 0.014 & 0.005 & 0.005 & \textbf{0.023}
   && 0.002 & \textbf{0.015} & 0.006 & 0.006 & \textbf{0.030} \\
\bottomrule
\end{tabular}%
}
\vspace*{-11pt}
\end{table}


Table~\ref{analogy_normal} shows the accuracy achieved by all models in the word analogy task, which is overall very poor. We attribute this to the size of the data set that contains less than the billions of tokens that are typically used to train for this task. GloVe performs better than word2vec for this task on both raw and annotated data, while VERSE does best overall.
The typed task, in which we also provide the entity type of the target word, is easier and results in better scores. If we consider only the subset of $6,892$ location targets for the GA task, we find that the graph-based models perform much better, with VERSE being able to predict up to $1,662$ $(24.1\%)$ of location targets on its best run, while $a$W2V and $a$GLV are only able to predict $14$ $(0.20\%)$ and $16$ $(0.23\%)$, respectively. For this entity-centric subtask, VERSE is clearly better suited. For the MSR task, which does not contain entities, we do not observe such an advantage.

The purity of clusters created with agglomerative clustering and mini-batch k-means for the categorization tasks are shown in Table~\ref{clustering_normal_p}, where the number of clusters were chosen based on the ground truth data. For the raw embeddings, we represent multi-word entities by the mean of the vectors of individual words in the entity's name. In most tasks, $r$W2V and $r$GLV create clusters with the best purity, even for the entity-based datasets of cities, politicians, and companies. However, most purity values lie in the range from $0.45$ to $0.65$ and no method performs exceptionally poorly. Since only the words in the evaluation datasets are clustered, the results do not give us insight into the spatial mixing of terms and entities. We consider this property in our visual exploration in Section~\ref{sec:exp}.

\begin{table}[t]
\caption{Categorization results. Shown is the purity of clusters obtained with k-means and agglomerative clustering (AC). The best two values are highlighted. For the raw text models, multi-word entity names are the mean of word vectors.}
\label{clustering_normal_p}
\setlength{\tabcolsep}{4pt} 
\renewcommand{\arraystretch}{1.0} 
\resizebox{\textwidth}{!}{%
\begin{tabular}{llccccccc}
\toprule
 && $r$W2V & $r$GLV & $a$W2V & $a$GLV & DW$_{id}$ & DW$_{log}$ & VRS \\
\midrule 
\multirow{6}{*}{k-means}  & ESSLLI\_1a & \textbf{0.575} & 0.545 & \textbf{0.593} & 0.454 & 0.570 & 0.520 & 0.534 \\ 
                          & ESSLLI\_2c & 0.455 & 0.462 & \textbf{0.522} & 0.464 & 0.471 & 0.480 & \textbf{0.584} \\
                          & Cities & \textbf{0.638} & \textbf{0.576} & 0.467 & 0.491 & 0.560 & 0.549 & 0.468 \\ 
                          & Politicians & \textbf{0.635} & 0.509 & 0.402 & 0.482 & 0.470 & 0.439 & \textbf{0.540} \\
                          & Companies & \textbf{0.697} & \textbf{0.566} & 0.505 & 0.487 & 0.504 & 0.534 & 0.540 \\
\cmidrule(lr){2-9}
                          & average & \textbf{0.600} & 0.532 & 0.498 & 0.476 & 0.515 & 0.504 & \textbf{0.533} \\ 
\midrule
\multirow{6}{*}{AC} & ESSLLI\_1a & 0.493 & \textbf{0.518} & 0.493 & 0.440 & 0.486 & 0.502 & \textbf{0.584} \\
                    & ESSLLI\_2c & 0.455 & 0.398 & 0.382 & 0.349 & \textbf{0.560} & \textbf{0.408} & 0.442 \\
                    & Cities & 0.447 & \textbf{0.580} & 0.440 & 0.515 & 0.364 & \textbf{0.549} & 0.359 \\
                    & Politicians & 0.477 & \textbf{0.510} & \textbf{0.482} & 0.480 & 0.355 & 0.360 & 0.355\\
                    & Companies & \textbf{0.511} & \textbf{0.519} & 0.475 & 0.504 & 0.474 & 0.469 & 0.473 \\
\cmidrule(lr){2-9}
                    & average & \textbf{0.477} & \textbf{0.505} & 0.454 & 0.458 & 0.448 & 0.458 & 0.443 \\ 
\bottomrule
\end{tabular}%
}
\vspace*{-12pt}
\end{table}

In summary, the results of the predominantly term-based intrinsic evaluation tasks indicate that a trivial embedding of words in an annotated corpus with state-of-the-art methods is possible and has acceptable performance, yet degrades the performance in comparison to a training on the raw corpus, and is thus not necessarily the best option. For tasks that include entities in general and require a measure of relatedness in particular, such as analogy task for entities or relatedness datasets, we find that the graph-based embeddings of VERSE often have a better performance. In the following, we thus explore the usefulness of the different embeddings for entity-centric tasks.

%% file: section_5_experimental.tex
Since there are no extrinsic evaluation tasks for entity embeddings, we cannot evaluate their performance on downstream tasks. We thus consider entity clustering and the neighborhood of entities to obtain an impression of the benefits that entity embeddings can offer over word embeddings for entity-centric tasks.

\begin{figure}[t]
\centering
\subcaptionbox{$r$W2V\label{sfig:word2vec_r_tsne}}{
\includegraphics[width=0.23\linewidth , height=0.21\linewidth]{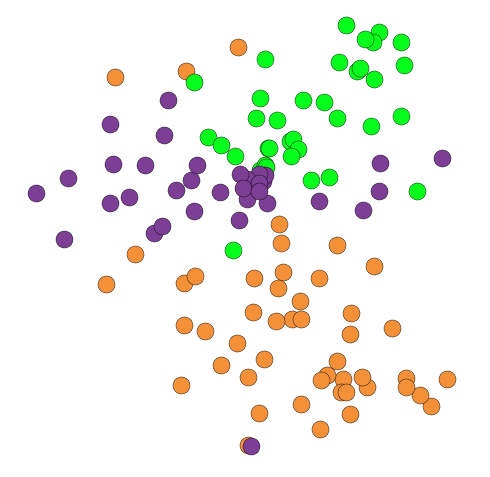}}
\subcaptionbox{$r$GLV\label{sfig:glove_r_tsne}}{
\includegraphics[width=0.23\linewidth , height=0.21\linewidth]{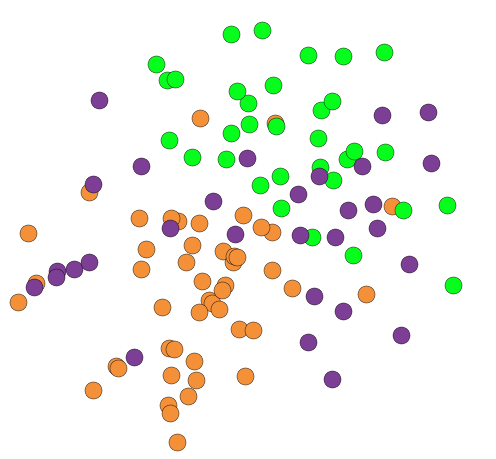}}
\subcaptionbox{$a$W2V\label{sfig:word2vec_tsne}}{
\includegraphics[width=0.23\linewidth , height=0.21\linewidth]{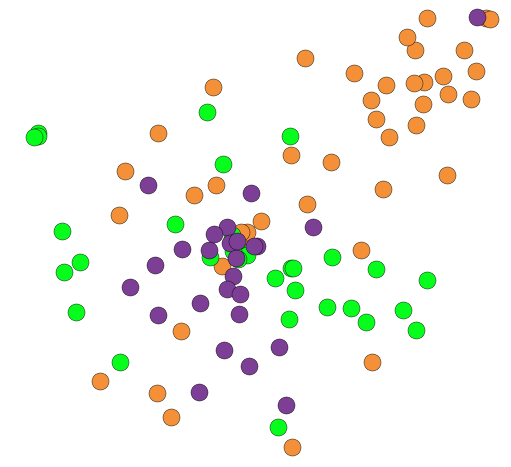}}
\subcaptionbox{$a$GLV\label{sfig:glove_tsne}}{
\includegraphics[width=0.23\linewidth , height=0.21\linewidth]{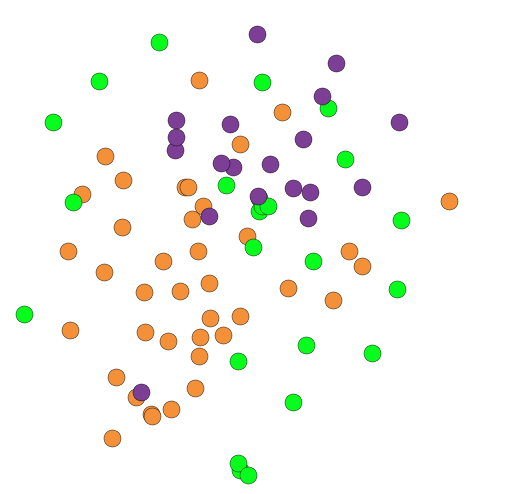}}
\subcaptionbox{DW$_{id}$\label{sfig:deepwalk_i_tsne}}{
\includegraphics[width=0.23\linewidth , height=0.21\linewidth]{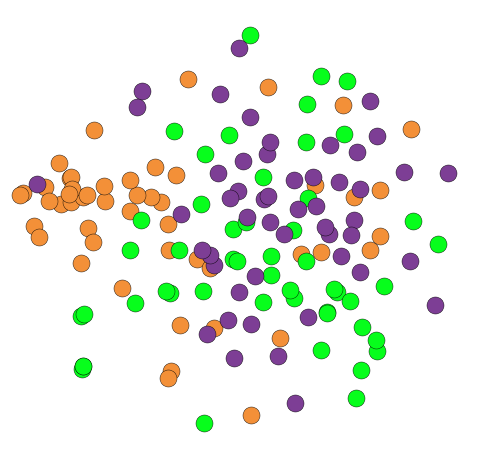}}
\subcaptionbox{DW$_{log}$\label{sfig:deepwalk_l_tsne}}{
\includegraphics[width=0.23\linewidth , height=0.21\linewidth]{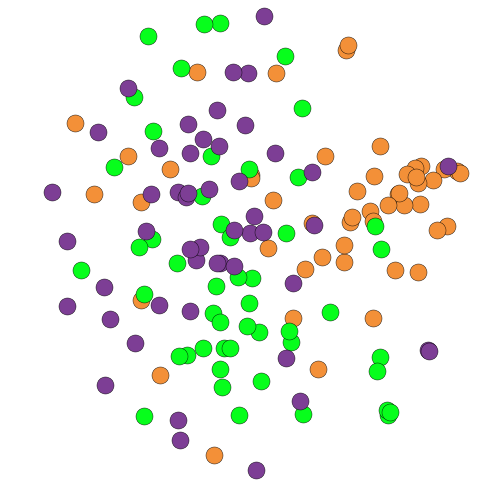}}
\subcaptionbox{VRS\label{sfig:verse_tsne}}{
\includegraphics[width=0.23\linewidth , height=0.21\linewidth]{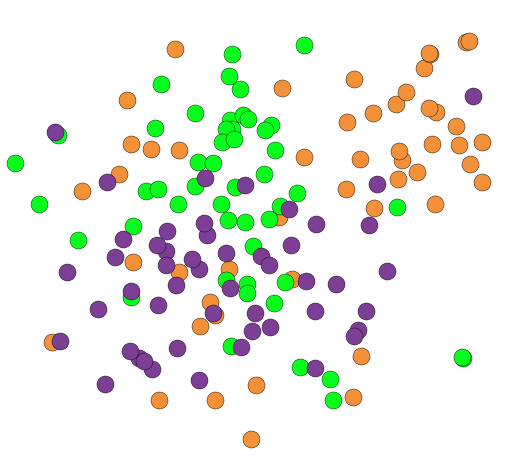}}
\vspace*{-5pt}
\caption{t-SNE projections of the embeddings for U.S.\ (purple), British (orange), and German (green) cities. For the raw text models, multi-word entity names are represented as the mean of word vectors.}
\label{fig:clust}
\vspace*{-8pt}
\end{figure}

\noindent
\textbf{Entity clustering.} 
To investigate how well the different methods support the clustering of similar entities, we consider 2-dimensional t-SNE projections of the embeddings of cities in Figure~\ref{fig:clust}. Since the training data is taken from news, we expect cities within a country to be spatially correlated. For the raw text embeddings, we represent cities with multi-component names as the average of the embeddings of their components. For this task, the word embeddings perform much better on the raw text than they do on the annotated text. However, the underlying assumption for the applicability of composite embeddings for multi-word entity names is the knowledge (or perfect recognition) of such entity names, which may not be available in practice.
The graph-based methods can recover some of the performance, but as long as entity labels are known, e.g., from a gazetteer, raw text embeddings are clearly preferable.

\begin{figure}[t]
\centering
\subcaptionbox{$r$W2V\label{sfig:obama_w2v_r}}{
\includegraphics[width=0.30\linewidth , height=0.28\linewidth]{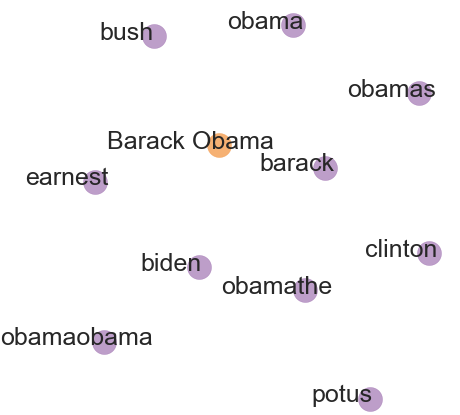}}\hspace{12pt}%
\subcaptionbox{$a$W2V\label{sfig:obama_w2v}}{
\includegraphics[width=0.30\linewidth , height=0.28\linewidth]{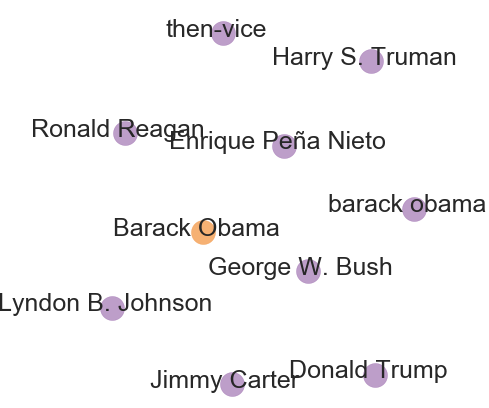}}\hspace{12pt}%
\subcaptionbox{VRS\label{sfig:obama_verse}}{
\includegraphics[width=0.30\linewidth , height=0.28\linewidth]{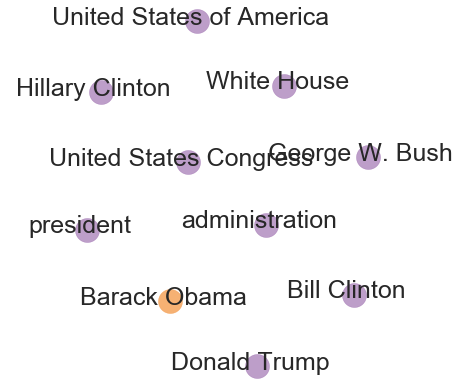}}
\vspace*{-5pt}
\caption{t-SNE projections of the nearest neighbours of entity \emph{Barack Obama}.
}
\label{fig:obama}
\vspace*{-15pt}
\end{figure}

\noindent
\textbf{Entity neighborhoods.} 
To better understand the proximity of embeddings, we consider the most similar neighbors by cosine similarity on the example of the entity \emph{Barack Obama}. Table~\ref{tbl:obama} contains a list of the four nearest neighbors of \emph{Barack Obama} for each embedding method. 
For the raw text models, we average the embeddings of the words \emph{barack} and \emph{obama}. Here, we find that the entity-centric models are more focused on entities, while the models that are trained on raw text put a stronger emphasis on terms in the neighborhood. In particular, $r$W2V performs very poorly and predominantly retrieves misspelled versions of the entity name. In contrast, even $a$W2V and $a$GLV retrieve more related entities, although the results for the graph-based embeddings are more informative of the input entity. Furthermore, we again observe a distinction in models between those that favor similarity and those that favor relatedness. 

The same trend is visible in the t-SNE projections of the nearest neighbours in Figure~\ref{fig:obama}, where word2vec primarily identifies synonymously used words on the raw corpus (i.e., variations of the entity name), and entities with an identical or similar role on the annotated corpus (i.e., other presidents). In contrast, VERSE identifies related entities with different roles, such as administrative locations, or the presidential candidates and the president-elect in the 2016 U.S.\ election.

\begin{table}[t]
\caption{Four nearest neighbours of entity \emph{Barack Obama} with cosine similarity scores. Entity types include terms T, persons P, and locations L.}
\label{tbl:obama}
\setlength{\tabcolsep}{2pt} 
\renewcommand{\arraystretch}{1.0} 
\resizebox{\textwidth}{!}{%
\begin{tabular}{lllllllllllllll}
\toprule
\multicolumn{3}{c}{$r$W2V} & & \multicolumn{3}{c}{$r$GLV} & & \multicolumn{3}{c}{$a$W2V} & & \multicolumn{3}{c}{$a$GLV} \\
\cmidrule(lr){1-3}
\cmidrule(lr){5-7}
\cmidrule(lr){9-11}
\cmidrule(lr){13-15}
T & obama      & 0.90 &&  T & obama          & 0.99 && P & George W. Bush         & 0.76 && T & president      & 0.78 \\
T & barack     & 0.74 &&  T & barack         & 0.98 && P & Jimmy Carter           & 0.73 && T & administration & 0.76 \\
T & obamaobama & 0.68 &&  T & president      & 0.77 && T & barack obama           & 0.73 && P & George W. Bush & 0.72 \\
T & obamathe   & 0.60 &&  T & administration & 0.74 && P & Enrique Pe{\~n}a Nieto & 0.67 && T & mr.            & 0.68 \\
\bottomrule         
\end{tabular}%
}

\hspace*{21pt}
\resizebox{0.87\textwidth}{!}{
\begin{tabular}{lllllllllll}
\toprule
\multicolumn{3}{c}{DW$_{id}$} & & \multicolumn{3}{c}{DW$_{log}$} & & \multicolumn{3}{c}{VRS} \\
\cmidrule(lr){1-3}
\cmidrule(lr){5-7}
\cmidrule(lr){9-11} 
L & White House    & 0.88 && L & White House    & 0.88 && L & White House              & 0.87 \\
T & president      & 0.79 && T & president      & 0.82 && T & president                & 0.79 \\ 
T & presidency     & 0.76 && P & George W. Bush & 0.78 && L & United States of America & 0.76 \\
T & administration & 0.75 && T & administration & 0.78 && P & Donald Trump             & 0.75 \\
\bottomrule
\end{tabular}%
}
\vspace*{-10pt}
\end{table}

%% file: section_6_conclusion.tex

We investigated the usefulness of vector embeddings of words in entity-annotated news texts. We considered the na\"{i}ve application of the popular models word2vec and GloVe to annotated texts, as well as node embeddings of cooccurrence graphs, and compared them to traditional word embeddings on a comprehensive set of term-focused evaluation tasks. Furthermore, we performed an entity-centric exploration of all embeddings to identify the strengths and weaknesses of each approach. 
While we found that word embeddings can be trained directly on annotated texts, they suffer from a degrading performance in traditional term-centric tasks, and often do poorly on tasks that require relatedness. In contrast, graph-based embeddings performed better for entity- and relatedness-centric tasks, but did worse for similarity-based tasks, and should thus not be used blindly in place of word embeddings.
Instead, we see potential applications of such entity embeddings in entity-centric tasks that benefit from relatedness relations instead of similarity relations, such as improved query expansion or learning to disambiguate,  which we consider to be the most promising future research directions and downstream tasks.

%% file: main_paper.bbl
\begin{thebibliography}{10}
\providecommand{\url}[1]{\texttt{#1}}
\providecommand{\urlprefix}{URL }

\bibitem{Abdi:2010:PCA:3160436.3160440}
Abdi, H., Williams, L.J.: {Principal Component Analysis}. {Wiley
  Interdisciplinary Reviews: Computational Statistics}  2(4),  433--459 (2010)

\bibitem{Agirre:2009:SSR:1620754.1620758}
Agirre, E., Alfonseca, E., Hall, K., Kravalova, J., Pa\c{s}ca, M., Soroa, A.:
  {A Study on Similarity and Relatedness Using Distributional and WordNet-based
  Approaches}. In: {Proceedings of Human Language Technologies: The 2009 Annual
  Conference of the North American Chapter of the Association for Computational
  Linguistics (NAACL-HLT)} (2009)

\bibitem{DBLP:conf/naacl/AgirreAHKPS09}
Agirre, E., Alfonseca, E., Hall, K.B., Kravalova, J., Pasca, M., Soroa, A.: {A
  Study on Similarity and Relatedness Using Distributional and WordNet-based
  Approaches}. In: {Human Language Technologies: Conference of the North
  American Chapter of the Association of Computational Linguistics (NAACL-HLT)}
  (2009)

\bibitem{DBLP:journals/corr/abs-1801-09536}
Bakarov, A.: {A Survey of Word Embeddings Evaluation Methods}  arxiv:1801.09536
  (2018)

\bibitem{Baroni}
Baroni, M., Evert, S., Lenci, A. (eds.): {Proceedings of the ESSLLI Workshop on
  Distributional Lexical Semantics Bridging the Gap Between Semantic Theory and
  Computational Simulations} (2008)

\bibitem{DBLP:conf/nips/BengioDV00}
Bengio, Y., Ducharme, R., Vincent, P.: {A Neural Probabilistic Language Model}.
  In: Advances in Neural Information Processing Systems (NIPS) (2000)

\bibitem{DBLP:journals/corr/BojanowskiGJM16}
Bojanowski, P., Grave, E., Joulin, A., Mikolov, T.: {Enriching Word Vectors
  with Subword Information}. {TACL}  5,  135--146 (2017)

\bibitem{Bruni:2014:MDS:2655713.2655714}
Bruni, E., Tran, N.K., Baroni, M.: {Multimodal Distributional Semantics}. J.
  Artif. Int. Res.  49(1),  1--47 (2014)

\bibitem{Das:2017:NER:3041821.3015467}
Das, A., Ganguly, D., Garain, U.: {Named Entity Recognition with Word
  Embeddings and Wikipedia Categories for a Low-Resource Language}. {ACM Trans.
  Asian {\&} Low-Resource Lang. Inf. Process.}  16(3) (2017)

\bibitem{Deerwester90indexingby}
Deerwester, S., Dumais, S.T., Furnas, G.W., Landauer, T.K., Harshman, R.:
  {Indexing by Latent Semantic Analysis}. {Journal of The American Society for
  Information Science}  41(6),  391--407 (1990)

\bibitem{DBLP:journals/corr/DiazMC16}
Diaz, F., Mitra, B., Craswell, N.: {Query Expansion with Locally-Trained Word
  Embeddings}. In: {Proceedings of the 54th Annual Meeting of the Association
  for Computational Linguistics (ACL), Volume 1: Long Papers} (2016)

\bibitem{DBLP:conf/eacl/DurmeRPM17}
Durme, B.V., Rastogi, P., Poliak, A., Martin, M.P.: {Efficient, Compositional,
  Order-sensitive n-gram Embeddings}. In: {Proceedings of the 15th Conference
  of the European Chapter of the Association for Computational Linguistics
  (EACL), Volume 2: Short Papers} (2017)

\bibitem{Ferret:2004:DWS:1220355.1220549}
Ferret, O.: {Discovering Word Senses from a Network of Lexical Cooccurrences}.
  In: {Proceedings of the 20th International Conference on Computational
  Linguistics (COLING)} (2004)

\bibitem{DBLP:journals/corr/GoldbergL14}
Goldberg, Y., Levy, O.: {Word2vec Explained: Deriving Mikolov et al.'s
  Negative-Sampling Word-Embedding Method}. {CoRR}  abs/1402.3722 (2014)

\bibitem{DBLP:journals/corr/GoyalF17}
Goyal, P., Ferrara, E.: {Graph Embedding Techniques, Applications, and
  Performance: A Survey}. {Knowl.-Based Syst.}  151,  78--94 (2018)

\bibitem{Node2vec}
Grover, A., Leskovec, J.: {node2vec: Scalable Feature Learning for Networks}.
  In: {Proceedings of the 22nd ACM SIGKDD International Conference on Knowledge
  Discovery and Data Mining (KDD)} (2016)

\bibitem{DBLP:journals/tacl/HillCKB16}
Hill, F., Cho, K., Korhonen, A., Bengio, Y.: {Learning to Understand Phrases by
  Embedding the Dictionary}. {TACL}  4,  17--30 (2016)

\bibitem{DBLP:journals/corr/HillRK14}
Hill, F., Reichart, R., Korhonen, A.: {SimLex-999: Evaluating Semantic Models
  With (Genuine) Similarity Estimation}. {Computational Linguistics}  41(4),
  665--695 (2015)

\bibitem{Klob}
Kolb, P.: {Experiments on the Difference Between Semantic Similarity and
  Relatedness}. In: {Proceedings of the 17th Nordic Conference of Computational
  Linguistics, (NODALIDA)} (2009)

\bibitem{Kuzi:2016:QEU:2983323.2983876}
Kuzi, S., Shtok, A., Kurland, O.: {Query Expansion Using Word Embeddings}. In:
  {Proceedings of the 25th ACM International Conference on Information and
  Knowledge Management (CIKM)} (2016)

\bibitem{R17-1057}
Lenc, L., Kr{\'{a}}l, P.: {Word Embeddings for Multi-label Document
  Classification}. In: {Proceedings of the International Conference Recent
  Advances in Natural Language Processing (RANLP)} (2017)

\bibitem{NIPS2014_5477}
Levy, O., Goldberg, Y.: {Neural Word Embedding as Implicit Matrix
  Factorization}. In: {Advances in Neural Information Processing Systems 27:
  Annual Conference on Neural Information Processing Systems (NIPS)} (2014)

\bibitem{W13-3512}
Luong, T., Socher, R., Manning, C.D.: {Better Word Representations with
  Recursive Neural Networks for Morphology}. In: Proceedings of the Seventeenth
  Conference on Computational Natural Language Learning (CoNLL) (2013)

\bibitem{vanDerMaaten2008}
Maaten, L.v.d., Hinton, G.: {Visualizing Data Using t-SNE}. {Journal of Machine
  Learning Research}  9(Nov),  2579--2605 (2008)

\bibitem{schutze2008introduction}
Manning, C.D., Raghavan, P., Sch{\"{u}}tze, H.: {Introduction to Information
  Retrieval}. Cambridge University Press (2008)

\bibitem{W04-3252}
Mihalcea, R., Tarau, P.: {TextRank: Bringing Order into Text}. In: {Proceedings
  of the 2004 Conference on Empirical Methods in Natural Language Processing
  (EMNLP)} (2004)

\bibitem{Mikolov}
Mikolov, T., Chen, K., Corrado, G., Dean, J.: {Efficient Estimation of Word
  Representations in Vector Space}. vol. arXiv:1301.3781 (2013)

\bibitem{mikolov2013linguistic}
Mikolov, T., Yih, W., Zweig, G.: {Linguistic Regularities in Continuous Space
  Word Representations}. In: {Human Language Technologies: Conference of the
  North American Chapter of the Association of Computational Linguistics
  (NAACL-HLT)} (2013)

\bibitem{DBLP:journals/cogsci/MitchellL10}
Mitchell, J., Lapata, M.: {Composition in Distributional Models of Semantics}.
  {Cognitive Science}  34(8),  1388--1429 (2010)

\bibitem{DBLP:conf/esws/MorenoBBDLRTG17}
Moreno, J.G., Besan{\c{c}}on, R., Beaumont, R., D'hondt, E., Ligozat, A.,
  Rosset, S., Tannier, X., Grau, B.: {Combining Word and Entity Embeddings for
  Entity Linking}. In: The Semantic Web - 14th International Conference,
  {ESWC}, Proceedings, Part {I}. pp. 337--352 (2017)

\bibitem{Nadeau}
Nadeau, D., Sekine, S.: {A survey of Named Entity Recognition and
  Classification}. {Lingvisticae Investigationes}  30(1),  3--26 (2007)

\bibitem{DBLP:conf/emnlp/PenningtonSM14}
Pennington, J., Socher, R., Manning, C.D.: {Glove: Global Vectors for Word
  Representation}. In: {Proceedings of the 2014 Conference on Empirical Methods
  in Natural Language Processing (EMNLP)} (2014)

\bibitem{Perozzi:2014:DOL:2623330.2623732}
Perozzi, B., Al{-}Rfou, R., Skiena, S.: {DeepWalk: online learning of social
  representations}. In: {The 20th ACM SIGKDD International Conference on
  Knowledge Discovery and Data Mining (KDD)} (2014)

\bibitem{Radinsky:2011:WTC:1963405.1963455}
Radinsky, K., Agichtein, E., Gabrilovich, E., Markovitch, S.: {A Word at a
  Time: Computing Word Relatedness Using Temporal Semantic Analysis}. In:
  {Proceedings of the 20th International Conference on World Wide Web (WWW)}
  (2011)

\bibitem{Rubenstein:1965:CCS:365628.365657}
Rubenstein, H., Goodenough, J.B.: {Contextual Correlates of Synonymy}. {Commun.
  ACM}  8(10),  627--633 (1965)

\bibitem{schnabel2015evaluation}
Schnabel, T., Labutov, I., Mimno, D.M., Joachims, T.: {Evaluation Methods for
  Unsupervised Word Embeddings}. In: {Proceedings of the 2015 Conference on
  Empirical Methods in Natural Language Processing (EMNLP)} (2015)

\bibitem{Spitz2016a}
Spitz, A., Gertz, M.: {Terms over {LOAD:} Leveraging Named Entities for
  Cross-Document Extraction and Summarization of Events}. In: {Proceedings of
  the 39th International {ACM} {SIGIR} Conference on Research and Development
  in Information Retrieval (SIGIR)} (2016)

\bibitem{DBLP:conf/ecir/SpitzG18}
Spitz, A., Gertz, M.: {Entity-Centric Topic Extraction and Exploration: {A}
  Network-Based Approach}. In: {Advances in Information Retrieval - 40th
  European Conference on {IR} Research (ECIR)} (2018)

\bibitem{DBLP:journals/lre/StrotgenG13}
Str{\"{o}}tgen, J., Gertz, M.: {Multilingual and Cross-domain Temporal
  Tagging}. {Language Resources and Evaluation}  47(2),  269--298 (2013)

\bibitem{tang2015line}
Tang, J., Qu, M., Wang, M., Zhang, M., Yan, J., Mei, Q.: {LINE: Large-scale
  Information Network Embedding}. In: {Proceedings of the 24th International
  Conference on World Wide Web (WWW)} (2015)

\bibitem{DBLP:conf/emnlp/ToutanovaCPPCG15}
Toutanova, K., Chen, D., Pantel, P., Poon, H., Choudhury, P., Gamon, M.:
  {Representing Text for Joint Embedding of Text and Knowledge Bases}. In:
  Proceedings of the 2015 Conference on Empirical Methods in Natural Language
  Processing, {EMNLP}. pp. 1499--1509 (2015)

\bibitem{DBLP:conf/naacl/ToutanovaKMS03}
Toutanova, K., Klein, D., Manning, C.D., Singer, Y.: {Feature-Rich
  Part-of-Speech Tagging with a Cyclic Dependency Network}. In: {Human Language
  Technology Conference of the North American Chapter of the Association for
  Computational Linguistics (HLT-NAACL)} (2003)

\bibitem{VERSE}
Tsitsulin, A., Mottin, D., Karras, P., M{\"{u}}ller, E.: {VERSE: Versatile
  Graph Embeddings from Similarity Measures}. In: {Proceedings of the 2018
  World Wide Web Conference on World Wide Web (WWW)} (2018)

\bibitem{DBLP:conf/emnlp/WangZFC14}
Wang, Z., Zhang, J., Feng, J., Chen, Z.: Knowledge graph and text jointly
  embedding. In: Proceedings of the 2014 Conference on Empirical Methods in
  Natural Language Processing, {EMNLP} , {A} meeting of SIGDAT, a Special
  Interest Group of the {ACL}. pp. 1591--1601 (2014)

\bibitem{DBLP:conf/conll/YamadaS0T16}
Yamada, I., Shindo, H., Takeda, H., Takefuji, Y.: {Joint Learning of the
  Embedding of Words and Entities for Named Entity Disambiguation}. In:
  Proceedings of the 20th {SIGNLL} Conference on Computational Natural Language
  Learning, CoNLL. pp. 250--259 (2016)

\bibitem{DBLP:conf/acl/YinS14}
Yin, W., Sch{\"{u}}tze, H.: {An Exploration of Embeddings for Generalized
  Phrases}. In: {Proceedings of the 52nd Annual Meeting of the Association for
  Computational Linguistics (ACL)} (2014)

\bibitem{DBLP:journals/corr/abs-1805-00280}
Zhou, D., Niu, S., Chen, S.: {Efficient Graph Computation for Node2Vec}. {CoRR}
   abs/1805.00280 (2018)

\end{thebibliography}
